\documentclass[sigconf, screen]{acmart}
\usepackage{subcaption}
\usepackage{booktabs}

\AtBeginDocument{%
  \providecommand\BibTeX{{%
    \normalfont B\kern-0.5em{\scshape i\kern-0.25em b}\kern-0.8em\TeX}}}

\copyrightyear{2021}
\acmYear{2021}
\setcopyright{acmlicensed}\acmConference[FDG'21]{The 16th International Conference on the Foundations of Digital Games (FDG) 2021}{August 3--6, 2021}{Montreal, QC, Canada}
\acmBooktitle{The 16th International Conference on the Foundations of Digital Games (FDG) 2021 (FDG'21), August 3--6, 2021, Montreal, QC, Canada}
\acmPrice{15.00}
\acmDOI{10.1145/3472538.3472600}
\acmISBN{978-1-4503-8422-3/21/08}

\begin{document}

\title{Ensemble Learning For Mega Man Level Generation}


\author{Bowei Li}
\affiliation{%
  \institution{University of Alberta}
  \streetaddress{Anon Address}
  \city{Edmonton}
  \country{Canada}}
\email{bowei4@ualberta.ca}

\author{Ruohan Chen}
\affiliation{%
  \institution{University of Alberta}
  \streetaddress{Anon Address}
  \city{Edmonton}
  \country{Canada}}
\email{ruohan2@ualberta.ca}

\author{Yuqing Xue}
\affiliation{%
  \institution{University of Alberta}
  \streetaddress{Anon Address}
  \city{Edmonton}
  \country{Canada}}
\email{yuqing6@ualberta.ca}

\author{Ricky Wang}
\affiliation{%
  \institution{University of Alberta}
  \streetaddress{Anon Address}
  \city{Edmonton}
  \country{Canada}}
\email{rwang4@ualberta.ca}

\author{Wenwen Li}
\affiliation{%
  \institution{University of Alberta}
  \streetaddress{Anon Address}
  \city{Edmonton}
  \country{Canada}}
\email{wenwen@ualberta.ca}

\author{Matthew Guzdial}
\affiliation{%
  \institution{University of Alberta}
  \streetaddress{Anon Address}
  \city{Edmonton}
  \country{Canada}}
\email{guzdial@ualberta.ca}

\renewcommand{\shortauthors}{Li et al.}

\begin{abstract}
Procedural content generation via machine learning (PCGML) is the process of procedurally generating game content using models trained on existing game content. 
PCGML methods can struggle to capture the true variance present in underlying data with a single model.
In this paper, we investigated the use of ensembles of Markov chains for procedurally generating \emph{Mega Man} levels.
We conduct an initial investigation of our approach and evaluate it on measures of playability and stylistic similarity in comparison to a non-ensemble, existing Markov chain approach.
\end{abstract}

\begin{CCSXML}
<ccs2012>
<concept>
<concept_id>10010147.10010257.10010321.10010333</concept_id>
<concept_desc>Computing methodologies~Ensemble methods</concept_desc>
<concept_significance>500</concept_significance>
</concept>
</ccs2012>
\end{CCSXML}

\ccsdesc[500]{Computing methodologies~Ensemble methods}

\keywords{procedural content generation, ensemble methods, markov chains, mega man}


\maketitle

%Novelty and Value

\section{Introduction}
Procedural content generation via machine learning (PCGML) is the process of procedurally generating game content using models trained on existing game content \cite{summerville2018procedural}. 
Although the usage of machine-learned models in PCGML can reduce the need for hand-coding design knowledge compared to traditional PCG \cite{summerville2018procedural}, it introduces new problems.
Many PCGML approaches use a single model that can struggle to capture the variance present in existing game content \cite{summerville2018expanding}. 
In particular, the high variance of game levels means that it may be better to model them with an ensemble of multiple models. 
However, popular PCGML approaches like neural networks that require more training data may not be appropriate for ensemble methods \cite{polikar2012ensemble}.
PCGML methods already suffer from a lack of training data, and ensemble methods divide the available data even further. 
Furthermore, models like neural networks have issues such as interpretability and training time \cite{summerville2018procedural}, making them not always preferable.
We identify an unexplored area of PCGML research: the usage of an ensemble of multiple simpler, interpretable models. 
%We are working towards a solution to both of these problems, combining multiple simpler models to better learn different behaviors to generate game content with high stylistic similarity while making the process easy to monitor and understand.

The current, common solutions to dealing with high variance in PCGML still employ a single model.
When confronted with high variance many researchers focus on modeling a sub-problem, for example training a model to generate subsections of levels \cite{sarkar2020sequential,awiszus2020toad} or by employing a more abstract representation \cite{summerville2016vglc}.
Alternatively, some researchers have used fitness functions to attempt to guide the machine learned model's generation when the available training data does not allow for sufficient generalization alone \cite{volz2018evolving,torrado2020bootstrapping}.
Much of this prior work focuses on \emph{Super Mario Bros.}, which is itself a relatively simple platformer game, with levels that only ever progress left-to-right.
These existing solutions may not be appropriate if one seeks to model more complex levels, such as larger levels that progress along both the x and y dimensions.  
%Many prior PCGML approaches proposed solutions to the issues that models like neural networks typically have by using simpler models that require less training time and data such as probabilistic graphical models. 
%Many of these approaches, however, only use one single model for a specific layer in a level or for the whole game level. 
%These approaches have been proved to work well on games with linear levels such as \emph{Super Mario Bros.}, where the levels all follow a left-to-right direction. 
%However, levels that move in multiple dimensions represent the kind of high variance game content and make these approaches struggle to capture the variance. One may consider reducing the variance by removing certain game features, but the generated content becomes less stylistically similar to the original game content. 

In our work, we explore procedurally generating game levels using an ensemble of simple models.
%that combines simpler models with higher interpretability and lower training time and data needed. 
Ensemble Learning combines multiple learning algorithms/models so that the ensemble can better capture the distribution of high variance training data \cite{polikar2012ensemble}. 
For this initial exploration of the topic, we trained multiple Markov chains to learn the structure of levels of the platformer game \emph{Mega Man}.
% and to generate new levels using the trained models.
Typical ensemble learning approaches train each model on a random split of their data, but as an initial investigation we focus on a domain-dependent method meant to better capture the dynamics of \emph{Mega Man} levels.
We evaluate our approach by comparing it to an existing, non-ensemble Markov chain approach \cite{snodgrass2015hierarchical}.

\begin{figure*}[hbt!]
  \centering
  \includegraphics[scale=0.50]{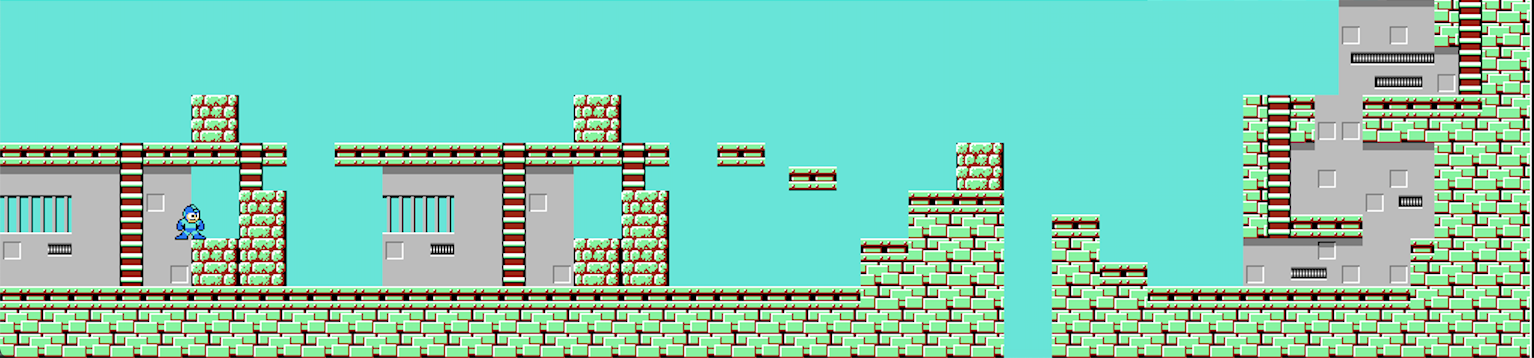}
  \caption{Section of a \emph{Mega Man} Level}
  \label{fig:2.1}
\end{figure*}

\begin{figure}[tb]
  \centering
  \includegraphics[scale=0.20]{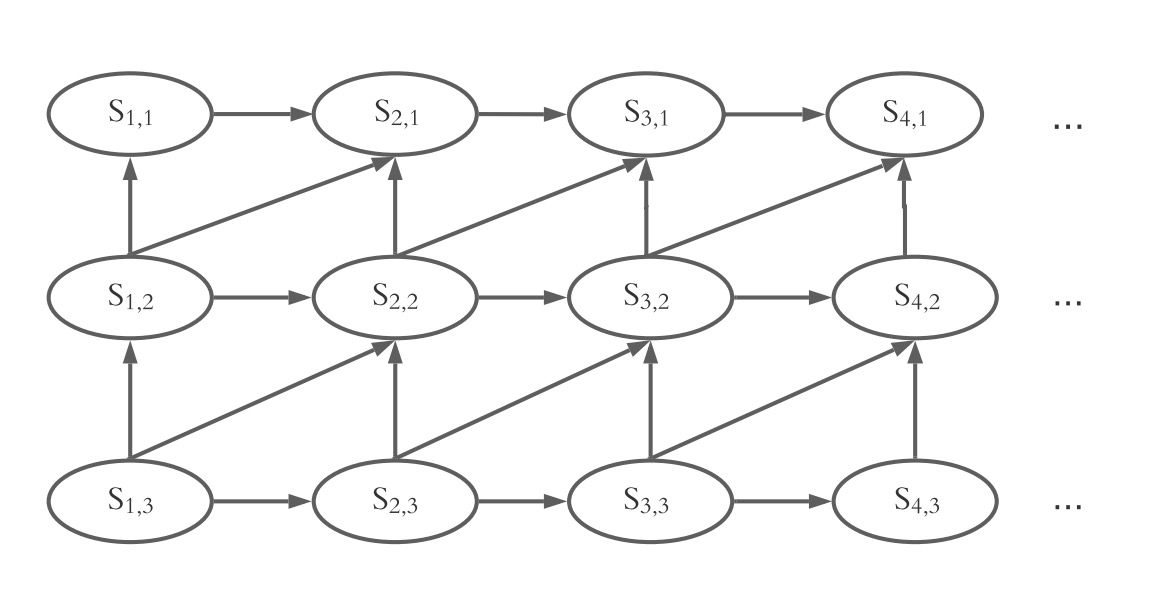}
  \caption{A Third Order Multidimensional Markov chain}
  \label{fig:2.2}
\end{figure}

The remainder of the paper is organized as follows.
First, we introduce the background on \emph{Mega Man} game levels as well as Markov chains. 
Second, we discuss related prior work. 
We then describe our approach for training our ensemble of Markov chains and how we generate new game levels from this ensemble. 
We evaluate the performance of our models on playability, the computational cost of generating new levels, and the stylistic similarity between the generated levels and the original game levels, all in comparison to an existing approach \cite{snodgrass2015hierarchical}. 
The paper closes with conclusions and directions for future work.

\section{Background}
In this section, we provide some background information about our game domain \emph{Mega Man} and Markov chains.

\subsection{\emph{Mega Man}}
\emph{Mega Man} is a platformer game series developed and released by Capcom \cite{kalata_2017}. 
In the original \emph{Mega Man}, the player takes on the role of \emph{Mega Man}, travelling across a land named Monsteropolis, fighting through enemies and performing various tasks in themed levels. 
Figure \ref{fig:2.1} shows a part of an original \emph{Mega Man} game level.
The goal of each \emph{Mega Man} level is to reach the end of the level, which requires platforming challenges and fighting various enemies. 
Notably, in comparison to other ``flat'' platformer levels like those from \emph{Super Mario Bros.}, \emph{Mega Man} levels require significantly more movement along the y-axis.

\subsection{Markov Chains}
A Markov chain is a machine learning method for modelling probabilistic transitions between states over time. 
A Markov chain is defined as a set of states \(S = \{s_1, s_2, ..., s_n\}\), and the conditional probability distribution (CPD) \(P(S_t|S_{t-1})\), representing the probability of transitioning to a state \(S_{t} \subseteq S\) given that the previous state was \(S_{t-1} \subseteq S\) \cite{markov1971extension}. 
Standard Markov chains restrict the probability distribution to only take the previous state into account. 
Higher-order Markov chains relax this condition by taking into account $k$ prior states, where $k$ is a finite natural number \cite{snodgrass2014experiments}. 
The CPD defining a Markov chain of order $k$ can be written as: \(P(S_t|S_{t-1}, ..., S_{t-k})\). 
That is, $P$ is the conditional probability of transitioning to a state $S_t$, given the states of the Markov chain in the past $k$ states.
 
A Multidimensional Markov chain (MdMC) is another extension of higher-order Markov chains that expands the structure of the model to take into account the surrounding states in a multidimensional graph instead of the past $k$ states as a sequence. 
In this way, the model can capture relations and dependencies more easily from 2D training data such as video game levels. 
In this work, we base our models on a third-order MdMC structure introduced by Snodgrass and Ontañón due to its simplicity \cite{snodgrass2013generating}, which we visualize in Figure \ref{fig:2.2}. 
The CPD defining this particular MdMC can be written as \(P(S_{t,r}|S_{t-1,r}, S_{t,r-1}, S_{t-1,r-1})\), where each state depends on the states to the left, below, and to the left and below in a 2D grid.

\section{Related Work}
In this section, we introduce prior work on Procedural Content Generation via Machine Learning (PCGML), focused on approaches employing probabilistic graphical models and ensemble methods.
%, such as Bayesian Networks \cite{guzdial2016toward, Summerville2015SamplingHS, Summerville2015TheLO}, Markov chains \cite{snodgrass2013generating, snodgrass2014experiments, snodgrass2014hierarchical, snodgrass2015hierarchical, snodgrass2016controllable, Summerville2015MCMCTSP4, 8974310}, and so on, along with the relationship between these works and ours.

Procedural Content Generation via Machine Learning (PCGML) is the generation of game content by machine-learned models that have been trained on existing game content \cite{summerville2018procedural}. 
%Many existing work utilize PCGML to generate video game content such as game levels from existing game content. 
There are a large number of PCGML approaches such as using n-grams \cite{dahlskog2014linear}, autoencoders \cite{jain2016autoencoders}, Long Short-Term Memory recurrent neural networks (LSTMs) \cite{summerville2016super}, reinforcement learning \cite{khalifa2020pcgrl}, and so on.
In this paper we focus on probabilistic graphical models, such as Markov chains and Bayesian Networks.  
Guzdial and Riedl \cite{guzdial2016toward} trained a Bayesian Network to generate \emph{Super Mario Bros.} levels from gameplay videos. 
Summerville et al. \cite{Summerville2015TheLO} investigated the use of a data-driven level generation approach using Bayesian Networks on Zelda dungeons.
We employ a Markov Chain over a Bayesian Network, as they represent simpler models that require less training data, and so are more suited to ensemble methods.
%These prior applications of Bayesian Networks they used learn observed variables as well as latent or hidden variables representing design rules, and are different types of probabilistic models from the Markov chains we used.

%Many prior PCGML work utilized Markov chains to generate game content. 
Markov chains have been a common method for PCGML since its inception \cite{snodgrass2013generating}.
Much of this prior work has focused on \emph{Super Mario Bros.} level generation, a common area of PCGML research \cite{snodgrass2013generating,snodgrass2014hierarchical,Summerville2015MCMCTSP4,snodgrass2015hierarchical}.
However, Markov chains have also been applied to games outside of \emph{Super Mario Bros.} and outside the platformer genre \cite{8974310}.
Of particular relevance to our paper is the work of Snodgrass and Ontañón, who first applied Markov chains to PCG \cite{snodgrass2013generating}.
They trained hierarchical Markov chains on \emph{Super Mario Bros.} levels, on both low-level tiles and high-level tiles \cite{snodgrass2014hierarchical,snodgrass2015hierarchical}. 
Although this approach performed well on games with linear levels such as \emph{Super Mario Bros.}, the authors stated it needed to be expanded to be able to handle games with nonlinear levels that progress in multiple directions such as \emph{Mega Man} or \emph{Metroid} \cite{snodgrass2014hierarchical}.
Our work was partially inspired by theirs, and we also employed a resampling technique similar to theirs \cite{snodgrass2016controllable} to process the content generated by our model. 
Our approach differs from theirs in that we use an ensemble of different Markov chains based on the concept of game paths to better capture both horizontal and vertical patterns. 
Snodgrass and Ontañón developed multi-layered representation of game levels to generate levels for \emph{Super Mario Bros.}, which employed multidimensional Markov chains \cite{snodgrass2017procedural}. 
They took the player path into consideration in this work.
We employ the concept of a ``game path'', which is distinct, as the player path describes the movement of the player through the level, whereas our game path represents the way a level itself moves along the horizontal and vertical axes.

\emph{Mega Man} has had significantly fewer PCGML approaches applied to it in comparison to \emph{Super Mario Bros.}.
Some prior work has modeled \emph{Mega Man} with machine learning without generating new \emph{Mega Man} content \cite{osborn2017automatic,guzdial2018automated}.
Sarkar et al. have modeled \emph{Mega Man} levels along with levels from a large number of other games with Variational Autoencoders (VAEs) for the purpose of recombining this content to create entirely new types of content \cite{sarkar2020sequential, sarkar2020conditional,yang2020game,sarkar2020exploring,sarkar2021generating}.
We instead focus on the problem of generating levels that resemble those from the original \emph{Mega Man}.
Recently, Capps and Schrum employed Generative Adversarial Networks (GANs) to model \emph{Mega Man} levels and modeled the direction of subsections of levels in a similar manner to our approach \cite{capps2021using}.
However, their output levels did not resemble the original \emph{Mega Man} levels in terms of shape, and the GANs they used were much more complex models than ours.
We demonstrate that our approach closely models original, unseen \emph{Mega Man} content, and outperforms an existing PCGML platformer level generation model, all with a much simpler approach.
%Sarkar and Cooper also covered \emph{Mega Man} level generation in their works \cite{sarkar2020sequential, sarkar2020conditional}, but they used Variational Autoencoders (VAEs) as their models.

To the best of our knowledge, the only prior PCGML work that has incorporated ensemble learning has been work that employs a Random Forest (RF).
However, all prior instances that have employed RFs have done so for a secondary classification task, and not for the primary generation task \cite{guzdial2018explainable, sarkar2020conditional, sarkar2020sequential}.
As such, our work stands out as the first PCGML approach to employ an ensemble of simple models for the primary generation task.

%There are many prior works utilizing ensemble learning that combines multiple learning algorithms/models to tackle their problems. 
%Summerville and Mateas \cite{Summerville2015SamplingHS} combined Bayes Networks and Principal Component Analysis to generate playable level graphs for the game \emph{The Legend of Zelda}. 
%Pereira et al. \cite{pereira2016learning} combined classifiers and an evolutionary approach to generate stable and feasible levels for the game \emph{Angry Birds}. 
%Raffe et al. \cite{raffe2014integrated} combined two integrated evolutionary searching cycles and a recommender system for player modeling to generate maps for the single-player action-shooter game \emph{Angry Bots}. 
%However, our work focuses on combining different Markov chains for the 2D platformer game \emph{Mega Man}. Therefore, our work differs from these in both the application domain and the models we use in the ensemble.

\begin{figure*}[hbt!]
  \centering
  \includegraphics[scale=0.30]{2-2megaman.png}
    \includegraphics[scale=0.14]{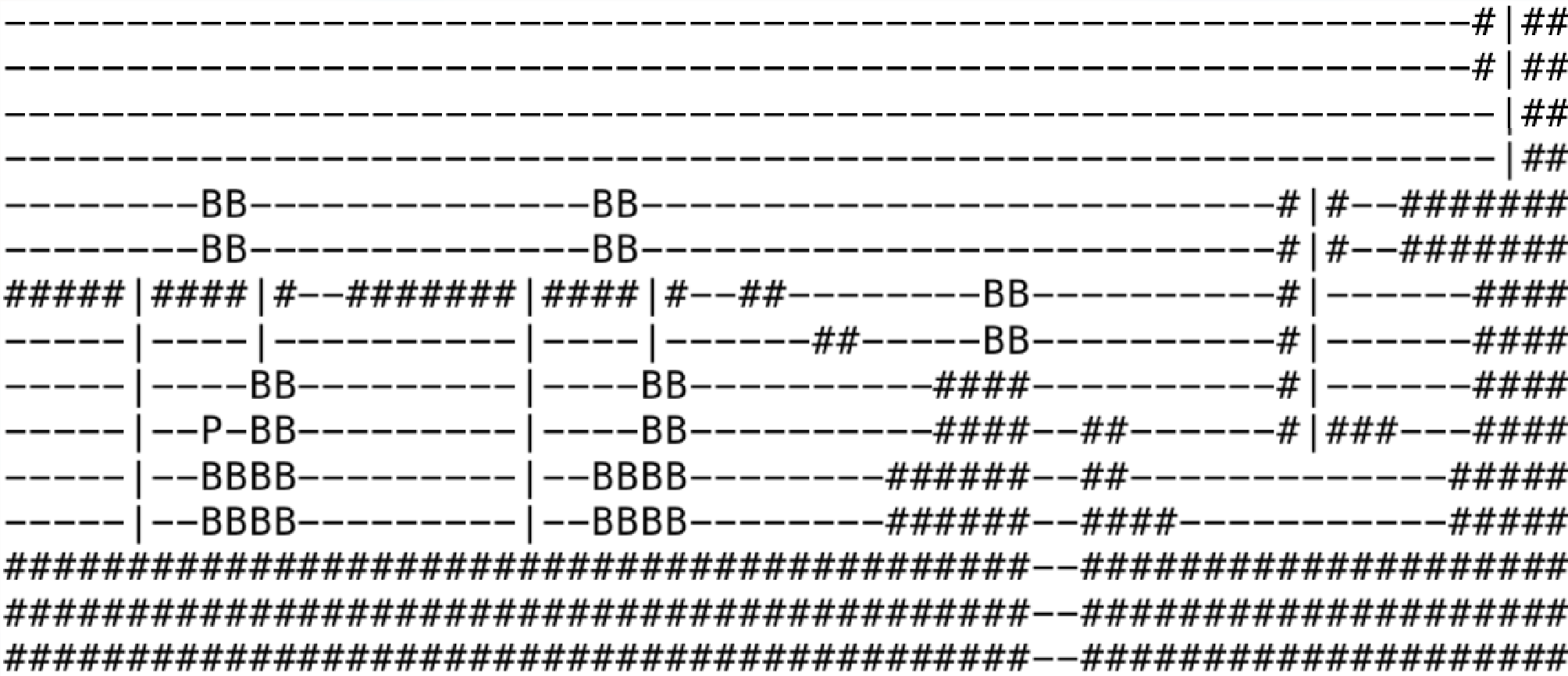}
  \caption{Left: An Original \emph{Mega Man} Level, Right: The Level Converted To Text Format From VGLC \cite{summerville2016vglc}}
  \label{fig:2}
\end{figure*}

\begin{figure*}
  \centering
  \includegraphics[scale=0.45]{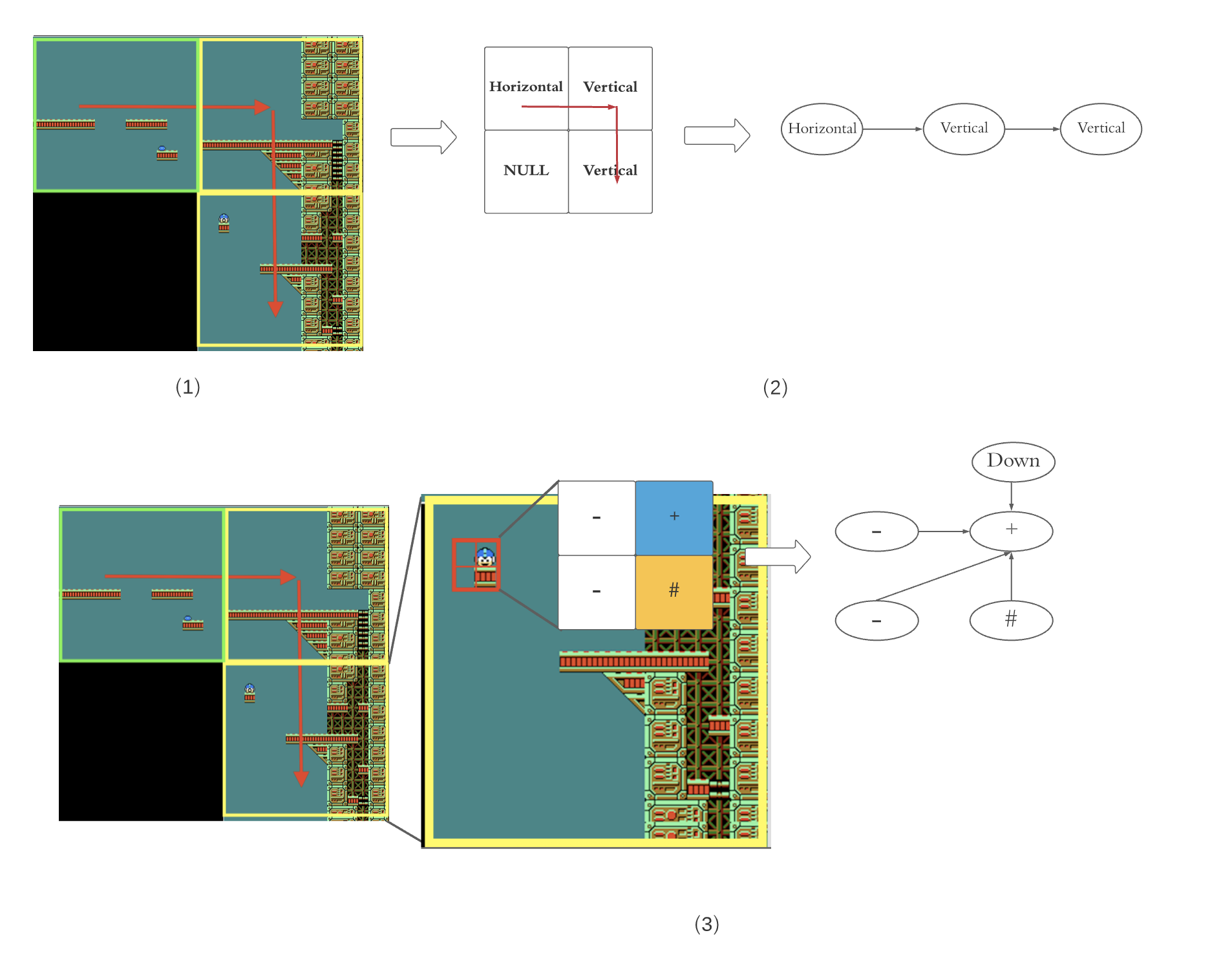}
  \caption{(1): Categorizing rooms. Yellow outline: vertical room; Green outline: horizontal room; Black: non-playable room. Red arrow: game path. (2): Learning high-level structure using simple Markov chains for the game path/room sequence. (3): Learning low-level structure within the rooms using room-specific Markov chains.}
  \label{fig:3}
\end{figure*}

\section{Approach}
In this section, we introduce our ensemble learning approach to train Markov chains to generate levels for the game \emph{Mega Man}.
Ensemble Learning combines multiple, simple learning models to achieve better performance and accuracy than could be obtained from a single model alone \cite{polikar2012ensemble}. 
Typically, one might randomly divide the training data into different buckets for each constituent model. 
In this case, as an initial exploration of this approach, we split the training data based on domain-specific knowledge. 
We employed three models in our approach. 
The first one is a first-order Markov chain for learning the high-level structure of the levels, which we call the ``game path''. 
To model the low level content of levels we divided all the existing levels into fixed-sized rooms and assigned them one of two types: horizontal and vertical.
Our second and third models are multidimensional Markov chains, one for the horizontal room type and one for the vertical room type.
%And then by dividing levels into fixed-sized rooms and categorizing them into two types, we trained one multidimensional Markov chain for the horizontal room type and one modified multidimensional Markov chain for the vertical room type.
We combine the rooms generated by our multidimensional Markov chains according to the high-level structure generated by the simple Markov chain to form a generated level. 
Our goal is to train models that can generate high-quality \emph{Mega Man} levels, and we hypothesize that we can achieve higher quality output levels with this approach, as opposed to attempting to model the problem with a single model.
This may seem counterintuitive, as with more data a machine learning model will often generalize more effectively. 
However, in the case of \emph{Mega Man}, which includes very different level structure in vertical or horizontal rooms, we anticipate that ensemble methods can be beneficial.

\subsection{Data}
Our training data consists of the 10 \emph{Mega Man} game levels from the Video Game Level Corpus (VGLC) \cite{summerville2016vglc}. 
The VGLC is an online collection of video game levels represented in an easy to parse format for PCGML and other game AI research purposes \cite{summerville2016vglc}. 
Figure \ref{fig:2} shows a part of an original game level from \emph{Mega Man} and the corresponding input text data after conversion. 
Each symbol in our input text file corresponds to a tile in the original level. 
For instance, the above figure shows that the symbol `\#' corresponds to a solid tile, a `l' corresponds to a ladder tile, a `-' corresponds to an empty tile, and so on.  
There are in total 18 different types of such tiles in \emph{Mega Man}, for full details please see the original VGLC paper \cite{summerville2016vglc}.

\subsection{Categorizing Rooms}

Unlike \emph{Super Mario Bros.}, games with levels like \emph{Mega Man} require players to progress both horizontally and vertically \cite{10.5555/3023108.3023128}.
To differentiate between different subsections of levels that require the player to progress in different dimensions, we employ the concept of rooms. 
We identify two types, horizontal rooms and vertical rooms, to represent these sections based on their features. 
In order to travel through a level in \emph{Mega Man}, the player needs to follow a fixed path of rooms.
We can thus identify how the player is expected to enter and exit these rooms. 
%Here by a fixed path we mean a fixed sequence of rooms.
We refer to each room with a left entrance and right exit as a horizontal room, and rooms with at least one entrance or exit in the up or down direction as a vertical room. 
In \emph{Mega Man}, the player begins in one of the leftmost rooms and ends up in one of the rightmost rooms, thus horizontal rooms always contain path segments going from left to right, not right to left. 

Our training process first converts a given level into rooms to capture the high-level structure of the level. 
We divide \emph{Mega Man} levels into 16*15 tile chunks, where 16 is the width, and 15 is the height of the chunk. We choose this specific size because most Mega Man levels from the VGLC can be divided into 16*15 tile chunks evenly.
Since we have the sequence of rooms for each input level, we are then able to label each chunk as a horizontal (H) or vertical (V) room. 
The picture at the top left corner in Figure \ref{fig:3} shows an example of such a conversion, where the red arrow denotes the game path and the yellow and green outlines denote vertical and horizontal rooms respectively. Additionally, we define non-playable areas that the player cannot enter as null rooms. 
We do not model these rooms explicitly, but we indirectly model them as the regions the game path does not travel through. 

\subsection{Learning High-level Level Structure}

We require some way to model each level as a sequences of rooms for the purpose of generating new sequences.
We employ a simple, linear Markov chain to model these sequences. 
Since every level of \emph{Mega Man} allows only one valid sequence of rooms that the player must follow to complete it, we can extract a one-dimensional sequence for each level. 
The rooms are labelled as horizontal or vertical. 
Figure \ref{fig:3} shows an example of this process. 

We employ a simple Markov chain to learn the probabilistic distribution of room sequences over our training set of levels.
We define the Markov chain as the probability of a certain type of room in the level given the prior room type. 
The prior room here means the last room that the path for this level passed through before passing through the current room. 
The CPD is defined as \(P(Room_i | Room_{i-1})\), where \(Room_i\) represents the \(i^{th}\) room in the room sequence. 
The trained Markov chain can then generate new room sequences of given lengths.
Once we have a generated room sequence, our ensemble of models can generate the content of each individual room.

\subsection{Learning Room Structure}

\begin{figure}
  \centering
  \includegraphics[scale=0.4]{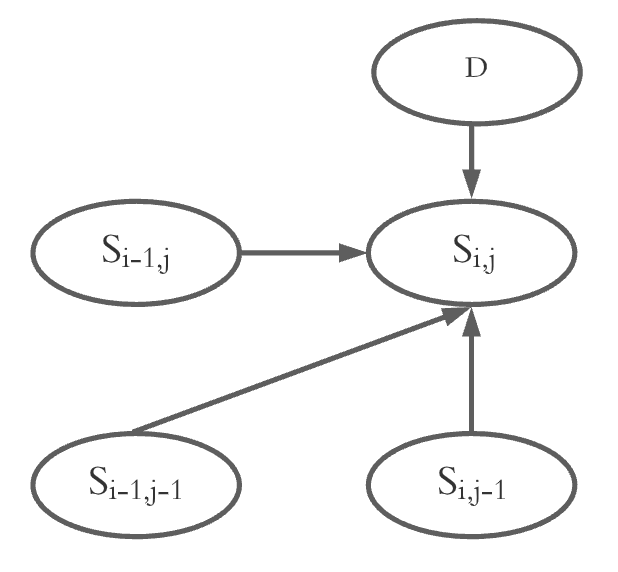}
  \caption{The visualization of the new CPD of \(S_{i,j}\), with D being the direction node}
  \label{fig:5}
\end{figure}

To model room structure we employed a variant of the multidimensional Markov chain from Snodgrass and Onta{\~n}{\'o}n \cite{snodgrass2014hierarchical}. 
They defined the nodes of the network to be the probability of a certain type of tile in the level given the tiles to its immediate left, bottom and bottom left.
In order to improve the performance of the L-shaped Markov chain in accurately learning the probabilistic distribution within a room, we trained two separate models, one for horizontal rooms and one for vertical rooms. 
This left us with an ensemble of these two models for the \emph{Mega Man} room generation task.

We  added an extra direction node to the original Snodgrass and Onta{\~n}{\'o}n model \cite{snodgrass2014hierarchical}. 
Unlike other nodes representing local tiles in the L-shaped Markov chain, the direction node represents the direction to the next room, which is also the direction that the player needs to travel in the current room. 
Figure \ref{fig:5} shows a visualization of the model that is defined by the CPD of the tile $s_{i,j}$, $P(s_{i,j} | s_{i-1,j}, s_{i,j-1}, s_{i-1,j-1}, D) $, where D is the direction node. 
Since the player always passes a horizontal room from left to right, the value of D is always ``right'' for horizontal rooms, indicating the next room is to the right of the current horizontal room. 
For vertical rooms, the value of D can be either ``up'', ``down'' or ``right'' depending on the relative position of the next room. Without the direction node, the model could fail to capture specific room structure.
For example in \emph{Mega Man}, the next room being above the current room usually leads to a higher probability of the room having climbable ladders, while the next room being below the current room does not, as the player can simply fall to move down. 
After training on the two different room types separately, we learned two modified multidimensional Markov chains that could each model the corresponding room behavior more accurately than training one such model on both room types, as we will demonstrate below.
Note that the use of the direction node can be seen as further dividing the model for learning vertical rooms into several sub-models that each learn a specific type of vertical rooms (e.g. the vertical rooms with path from top to bottom, from bottom to top, from left to top, and so on). 
Figure \ref{fig:3} shows an example of using such a modified L-shaped Markov chain to learn the low-level structure within a vertical room.

\subsection{Generation}

\begin{figure}
  \centering
  \includegraphics[scale=0.25]{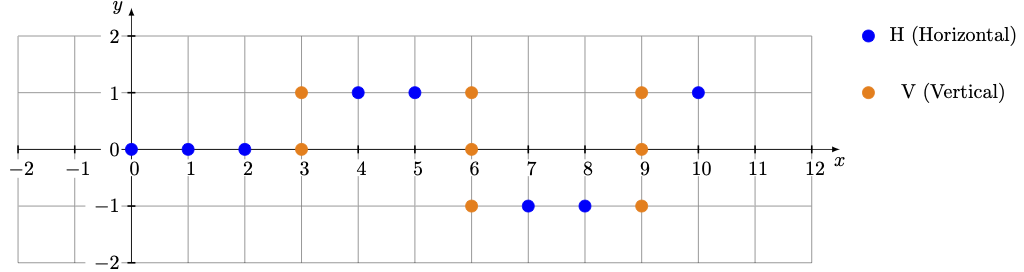}
    \includegraphics[scale=0.29]{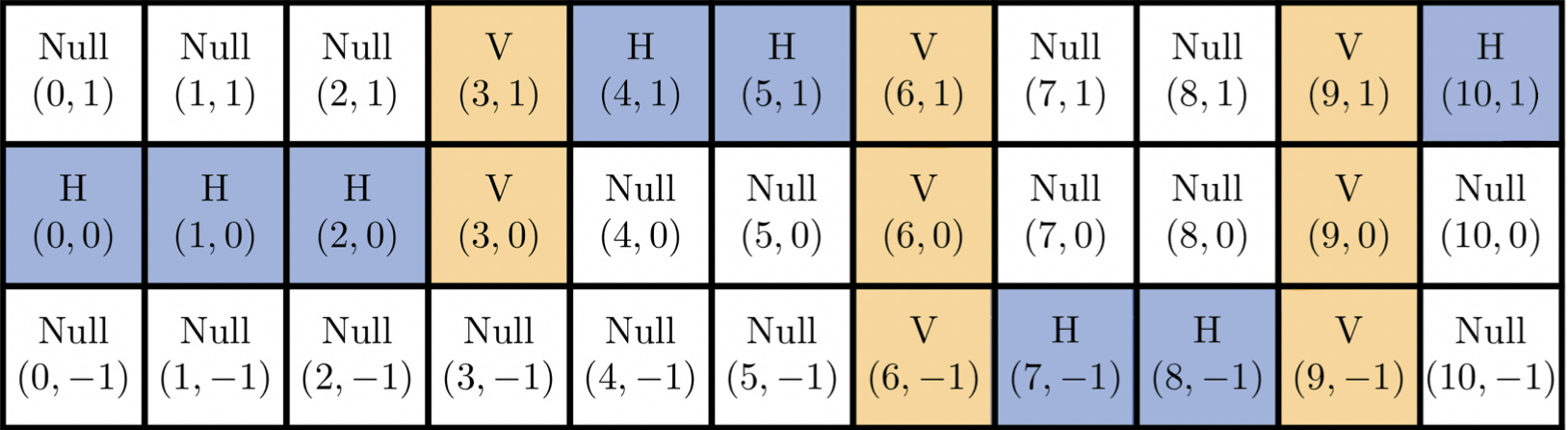}
  \caption{Top: The Generated Coordinate Plane; Bottom: The Generated Incomplete Level}
  \label{fig:6}
\end{figure}

When generating a level, we first produce a room sequence using our high-level model. 
After randomly choosing the type for the first room in the sequence (either horizontal or vertical), we use the simple Markov chain to probabilistically generate the next room given the previous one until we reach a sequence of a given length. 
After we produce the room sequence, we build the corresponding level structure.
Our model first creates a coordinate plane with the first room in the room sequence at the origin point. 
Every subsequent room in the room sequence is assigned a coordinate according to the room type and the locations of the adjacent rooms. 
For example, assuming that the first room in the room sequence is a vertical room at the origin point $(0, 0)$, a subsequent horizontal room will be plotted at $(1, 0)$, while a subsequent vertical room can be placed at either $(0, -1)$ or $(0, 1)$ depending on the randomly generated direction (moving upwards or downwards). 
The coordinate plane can then be converted to an empty level consisting of several, initially empty rooms. 
Any ``room'' not on the path is considered a null room, and is left blank.
Figure \ref{fig:6} shows the coordinate plane and an empty level corresponding to room sequence:

\begin{center}
HHHVVHHVVVHHVVVH
\end{center}\par

After producing the level structure, we then fill the empty rooms with content generated by our two modified L-Shaped Markov Chains based on their type. 
When generating rooms, our model generates a new room one tile at a time starting from the bottom left corner. It first generates the bottom row, followed by the row above it, and so on. 
It keeps generating rows until there are 15 rows (15 is the height of each room). 
Tiles that are generated by our method are selected probabilistically, based on the learned probability distribution. 
Note that while the process of generating a horizontal room is the same as that of generating a vertical room, the two generation processes are based on different probability distributions generated by two different L-shaped Markov chains. 
If we encounter a combination of previous tiles that were not seen during training, we put an empty tile in that position for simplicity. 
Rooms labelled as null are non-playable and are filled with null tiles. 

We integrate a resampling method that guarantees an accessible path between the border of adjacent rooms when generating a level. 
The method takes two rooms along with their relative positions as inputs. 
With the relative positions specified, our method checks if the joint between the two rooms is open. 
For example, if the next room is on the right side of the current room, our resampling method will determine whether the path that connects the rightmost column of the current room and the leftmost column of the next room is blocked or not. 
If it is blocked, the program will generate a new room until the path is no longer blocked.

\section{Evaluation}   

In this section, we describe the evaluation of our approach in comparison to two baselines. 
The evaluations are based on the following five metrics: layout playability, overall playability, necessary resampling, layout similarity, and inverse content similarity.
We describe these in more detail below.

\subsection{Experimental Setup}
For our experiments, we used nine levels from \emph{Mega Man} to train our models, with one level withheld as a test level. 
In order to evaluate each approach, we generated 50 levels with 12 playable rooms each for our approach and our two baselines.
We evaluated them using these metrics:

\begin{itemize}
    \item \emph{Layout Playability}: the percentage of playable room layouts or game paths. 
    Note that we are evaluating the high-level layout of rooms without considering the structure within each room when evaluating the layout playability. 
    Every room is viewed as a high-level tile in a layout. 
    If there exists a path that goes from the starting room to the ending room without any null rooms breaking the path, then the layout is playable. 
    In a room sequence, the starting room and the ending room are automatically defined as the first and last rooms in the sequence. 
    In a room layout generated by our baselines, we manually define the starting room and ending room as a leftmost and rightmost room respectively. 
    We choose the layout playability as one of our metrics because in our approach, the high-level structure and the low-level structure are learned separately using different models. 
    A higher layout playability indicates a better performance of our high-level model in generating coherent high-level structure of levels.
    \item \emph{Overall Playability}: the percentage of 50 generated levels that are playable. 
    To decide whether the whole generated level is playable, we implemented an A* agent specially designed for tile-based \emph{Mega Man} levels to find paths through the levels. We approximated the Mega Man-specific max jump height and width from the annotated player path information provided by the VGLC Mega Man level data \cite{summerville2016vglc}. This jump information was then used by our A* agent to simulate jumps in the tile representation.
    If the A* agent finds a path for a level, then we conclude that this level is playable. 
    However, the A* agent sometimes will fail to find a path even for some playable levels. 
    This happens under a few circumstances, such as the player encountering some unavoidable traps on the optimal path that reduce the player’s health, or the player needing to pass through a room with the help of some movable objects. 
    The A* agent will stop at the traps and end up failing to find another path, and since the level generated by our model is static, the A* agent cannot take advantage of the movable objects as in the original game.
    Therefore, the number of playable levels given by our A* agent should be understood as a lower bound.
     We choose to measure the overall playability of our generated levels because this could help us understand how our three Markov models perform at generating coherent levels in comparison to our baselines.
     %could give us a good idea of how our three models work together in generating coherent levels.
    \item \emph{Necessary Resampling}: We count the average number of times the generated levels for each level needed to resample rooms. 
    We prefer a lower number for this metric since resampling requires extra computation power, and a smaller number indicates a better performance of our models in generating coherent levels.
    \item \emph{Layout Similarity}: We measure the probabilistic similarity between room sequence model and the layout of the withheld test level. 
    Since the test level was one used in the original video game, we assume that its layout has suitable complexity and represents desired structure for generated levels. 
    We calculate the probability of our high-level model generating the layout of the test level to evaluate the similarity between the layouts. 
    The similarity between the layouts of our generated levels and the test level can be a good indicator of how complex the layouts of our generated levels are, so we prefer a higher value for this metric. 
    \item \emph{Inverse Content Similarity}: We measure the stylistic similarity between the content of rooms generated by our room-level models and the content of rooms from the test level we chose. 
    As above, we assume the rooms from the test level have suitable complexity and desired structure.
    The way we evaluate this similarity is as follows: 
    when we evaluate the inverse content similarity of our room-level models, we train a second ensemble using the rooms from the test level as the training data. 
    We then compare the conditional probabilistic distribution (CPD) of our existing models with the CPD of the new models that reflects the tile distribution patterns in the test level. 
    For example, when we evaluate an L-shaped model, we get the total difference between these two CPDs by measuring the sum of squared difference
    \[\sum_{i,j}(P( s_{i,j}|s_{i-1,j}, s_{i,j-1}, s_{i-1,j-1}) - P'( s_{i,j}|s_{i-1,j},  s_{i,j-1},  s_{i-1,j-1}))^2 \]
    where \(s_{i,j}\) denotes one type of tile, \(P(s_{i,j}|s_{i-1,j}, s_{i,j-1},s_{i-1,j-1})\) is the probability of generating a tile of the type \(s_{i,j}\) given three other tile types \((s_{i-1,j}, s_{i,j-1}, s_{i-1,j-1})\) using our L-shaped model, and \(P'\) is the probability distribution in the test level. 
    For each combination of three tile types \((s_{i-1,j}, s_{i,j-1}, s_{i-1,j-1})\) in the test level, we find all possible next tile types \(s_{i,j}\) and the corresponding conditional probabilities both in the test level and in our L-shaped models, and calculate the sum of squared differences between the probabilities. 
    After calculating the squared sum of differences, we also calculate the mean difference by dividing the total difference by the total number of entries in the conditional probability table. 
    Because we want the content generated by our model to be similar to the content of the test level, we prefer a lower difference for our models. 

\end{itemize}

We compare our approach with these two baselines:
\begin{enumerate}
    \item \emph{Simple Hierarchical Markov Chains (MCs)} This approach was previously introduced by Snodgrass and Ontañón \cite{snodgrass2014hierarchical}. 
    By setting it as a baseline and comparing our approach to it on the metrics we chose, we can determine whether using this single Markov chain is sufficient for achieving high quality output levels, and specifically how it compares to our ensemble approach.
    
    \item \emph{Our Simplified Approach} This baseline is a simplified version of our approach. 
    The difference here is that we employ a single L-shaped Markov chain with the directional node as seen in Figure \ref{fig:5}, trained on all available room data.
    %The difference here is whether we employ ensemble learning that combines two different L-shaped models to learn different room structures separately. 
    Since this baseline shares the same way of modeling the high-level, room layout structure with our approach, we can gain a more direct view of whether ensemble learning has an impact on the overall playability and stylistic similarity of the generated levels.
    One might naively assume this approach should outperform our ensemble, since it will have much more data to train on, and can therefore theoretically generalize more effectively.
    Thus it is an important baseline for demonstrating the benefits of our approach. 
\end{enumerate}

\section{Results}

\begin{table*}
  \caption{Results of the Three Approaches}
  \label{sample-table}
  \centering
  \begin{tabular}{llll}
    \toprule
                            & Simple Hierarchical MC  &  Our Simplified Approach    & Our Approach \\
    \midrule
    Layout Playability      & 62.0\%     & 96.0\%    & \textbf{98.0\%} \\
    Overall Playability     & 6.0\%      & 16.0\%    & \textbf{32.0\%} \\
    Necessary Resampling    & 58.2     & 10.4      & \textbf{6.6}    \\
    Layout Similarity       & <0.1\%  & \textbf{2.4\%} & \textbf{2.4\%} \\
    \bottomrule
  \end{tabular}
\end{table*}

\begin{table*}
  \caption{Comparison of Inverse Content Similarity of the Two Low-Level Models}
  \label{sample-table2}
  \centering
  \begin{tabular}{llll}
    \toprule
                            & Our Simplified Approach  & Our Approach\\
    \midrule
    Total Difference        & 0.867148      & \textbf{0.145836} \\
    Mean Difference         & 0.003496      & \textbf{0.000343} \\    
    \bottomrule
  \end{tabular}
\end{table*}

Table~\ref{sample-table} shows the average or percentile values of our metrics over the 50 generated levels from each approach. 
As we can see from the first row in the table, when generating layouts of levels, our approach outperforms the Simple Hierarchical MCs used by Snodgrass and Ontañón \cite{snodgrass2014hierarchical}. 
Only 62\% of the layouts generated by the L-shaped Markov chain are playable, since some misplaced null rooms (non-playable areas) often block the path from source room to destination room. 
On the other hand, over 95\% of the layouts generated by the simple Markov chain method used by our approach are playable. 
That is because the room sequences that we used to represent the layout only consist of playable rooms. 
The high-level structure constructed from the room sequences and our assembling approach is more likely to be playable compared to the structure generated by the L-shaped Markov chain. 
Notice that Our Approach, Our Simplified Approach both used the simple Markov chain for layout generation, so the difference between these two results in layout playability is due to random chance.

The second row in Table ~\ref{sample-table} shows the overall playability of the levels generated by the three approaches. 
As we can see, generating playable levels is much harder than generating playable layouts as we take the content of each room into consideration. 
Only 6\% of the levels generated by the Simple Hierarchical MCs are playable. 
The performance of this approach was once again affected by the misplaced null rooms, which led to levels with rooms that were disconnected from one another.
Our approach generates a higher number of playable levels compared to our simplified approach, which can only be attributed to our use of ensemble learning.
Despite Our Simplified Approach having more training data by training one model on all available rooms, it performed worse at generating coherent rooms. 
We expect this is due to the fact that horizontal and vertical rooms have significantly different content, which means that modeling them separately leads to better performance than attempting to generalize between them.
However, Our Simplified Approach also outperformed the Simple Hierarchical MC, which we attribute in part to the direction node.
For example, a room is more likely to have a higher number of climbable objects such as ladders if the player is supposed to travel up to proceed to the next room, i.e. if the next room is at the top of the current room. 
On the other hand, if the next room is below the current room, then the current room is less likely to have a row of spikes or traps at the bottom of the room.
By adding the direction node, our models are better able to capture these behaviors and therefore able to generate more playable levels. 

The third row in Table ~\ref{sample-table} shows the average number of room resamplings needed for generating a level for each of the three approaches. 
Note that the resampling was done before the playability check.
A high number indicates a model has less probability of generating coherent rooms, and we can clearly see that the Simple Hierarchical MC approach requires more resampling than the other two approaches. 
That is mainly because the misplaced null rooms in the layout generated by the L-shaped Markov chain often makes it impossible to proceed in the level, leading to continuous resampling. 
In this case, we had to include 70 as a maximum number of resampling, and just conclude that the levels were unplayable.
Our Approach again yields a better result than Our Simplified Approach, which indicates the benefits of ensemble learning to split up the task of room generation.

The last row in Table ~\ref{sample-table} shows the stylistic similarity between the level layouts generated by our high-level models and the layout of the test level we chose. 
Note that the Simple Hierarchical MC approach used the L-shaped Markov chain to model the entire tile layout, whereas Our Simplified Approach and Our Approach both used the higher-level, simple Markov chain to model the room layout structure. 
The L-shaped Markov chain has a very low chance (<0.1\%) of generating the same layout structure as the test level. 
In contrast, the simple Markov chain did a better job of learning and reproducing the same layout structure, having a chance of 2.4\% in generating the layout of the test level.
We attribute this low number to the large number of layouts that can be generated by our approach. 
However, the fact that it is above the Simple Hierarchical MC likely indicates that more of the generated layouts are similar to those from the original \emph{Mega Man} levels.

Table ~\ref{sample-table2} shows the stylistic similarity between the content within rooms from the test level and the content within rooms generated by Our Simplified Approach and Our Approach. 
As we mentioned above, we evaluate the inverse content similarity by measuring the difference between the conditional probability distribution (CPD) of our low-level models and the CPD of tiles in the test level. 
The total difference is the sum of squared differences between all entries from the two conditional probability tables. 
Each entry in the conditional probability table is the probability of observing a certain type of tile given three other surrounding tile types. 
We divide the total difference by the number of entries to get the mean difference between the two CPDs. 
The mean difference is therefore a good indicator of the difference between the CPD of our low-level models and the CPD of tiles in the test level.
When the difference is smaller, a learned model is more likely to generate the same content within each room as in the test level. 
We prefer smaller differences and therefore greater similarity to indicate our models have a good coverage of \emph{Mega Man} structure.
As we can see from the table, our approach is more capable of capturing room structure as the differences are smaller. 
%For example in \emph{Mega Man}, the next room being above the current room usually leads to a higher probability of having climbable ladders, while the next room being below the current room does not, as the player can simply jump off some objects to move downward. 
Thus we can confirm that training multiple models led to better coverage of the structures of different types of rooms in comparison to training one model on all available rooms.

\begin{figure*}
  \centering
  \includegraphics[scale=0.30]{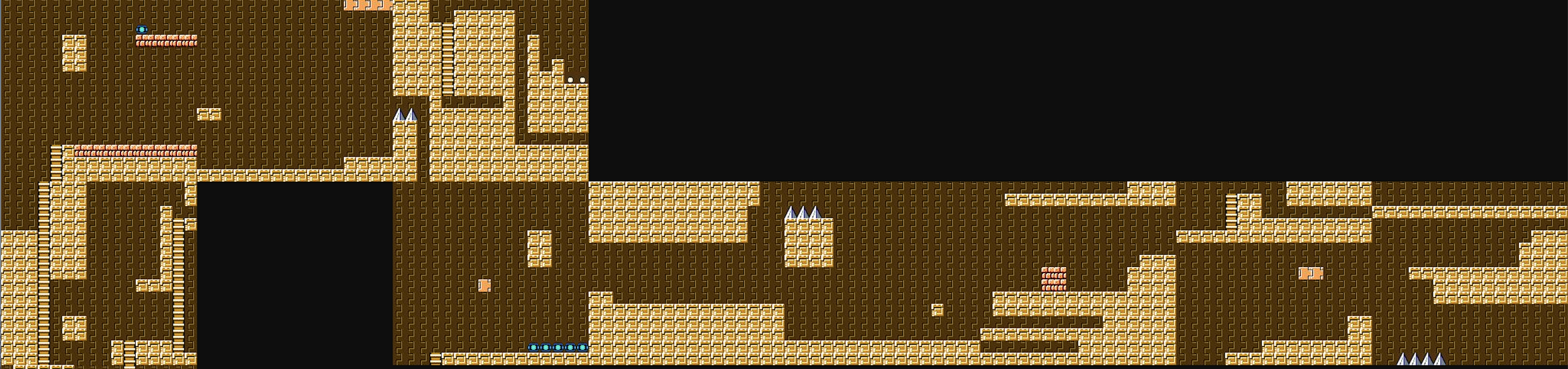}
  \caption{An output level generated by our approach}
  \label{fig:7}
\end{figure*}

\section{Discussion}

From the above observations, the simple Markov chain we used has outperformed the L-shaped Markov chain in learning and generating sequences of rooms, and our approach also outperformed our simplified approach in learning and generating complex room content that is similar to the content in the original game \emph{Mega Man}. 
Therefore, we can conclude that our approach did an overall better job than the other two baselines in producing content that is more stylistically similar to the original game levels. 
In Table ~\ref{sample-table}, we also observed that our approach has a higher probability of generating playable content with less cost from resampling than the other two baselines. 
We conclude that by using an ensemble of multiple models, our approach has done an overall better job in generating \emph{Mega Man} game levels. 
We would recommend employing an ensemble of models for PCGML tasks that involve modeling content with a high degree of variance, where that variance can be reduced by splitting up the available training data.

The base learners in our ensemble, which are the three different Markov chains, are all simple Markov chain models.
Snodgrass and Ontañón later introduced more complex multidimensional Markov chains \cite{snodgrass2017procedural} and Markov Random Fields \cite{snodgrass2017Learning} on level generation for \emph{Super Mario Bros.}, \emph{Kid Icarus}, and \emph{Loderunner}.
We avoided these more complex models as part of our initial investigation into ensemble learning for PCGML, but we hope to explore them in future work.

Figure \ref{fig:7} shows an output level from our approach. 
The player enters the level from the room at the bottom left corner and exits on the rightmost room of the level. 
This level was playable by our A* pathfinding agent.
However, we can see that it still has a number of issues.
For example, there is level content that the player would be cut off from, as in the fourth, upper rightmost room. 
while the level is not equivalent to a human-authored \emph{Mega Man} level, it does contain interesting structure, and employs more tile types than just the basic solid and empty background tiles, which we think compares favorably with prior work employing deep neural networks for generating \emph{Mega Man} levels \cite{capps2021using}.
%and contains different tile types that makes it more complex and fun to play in.

%In the future, we want to explore other methods in dividing and categorizing contents of game levels, as well as other ways of combining different models to efficiently generate game levels with a limited amount of data.

\section{Conclusion}
We explored the use of ensemble learning on video game level generation using a combination of multiple Markov chains. 
By dividing game levels into rooms and categorizing them using the concept of game paths, we were able to use a simple Markov chain to learn the high-level structure of levels and generate room sequences, leading to high layout playability. 
Based on the categorization, we trained multiple multidimensional Markov chains, one for each type of room, to better capture low-level structure of \emph{Mega Man} game levels.
The use of ensemble learning outperformed an existing Markov method, and a variation of our approach without ensemble learning.
We encourage future PCGML researchers to employ ensembles of models to better model high variance game content.

\begin{acks}
The authors would like to thank Dr. Russ Greiner for his guidance towards the creation of this paper. 
\end{acks}

\bibliographystyle{ACM-Reference-Format}
\bibliography{main}


\begin{thebibliography}{34}

%%% the Web address.

\ifx \showCODEN    \undefined \def \showCODEN     #1{\unskip}     \fi
\ifx \showDOI      \undefined \def \showDOI       #1{#1}\fi
\ifx \showISBNx    \undefined \def \showISBNx     #1{\unskip}     \fi
\ifx \showISBNxiii \undefined \def \showISBNxiii  #1{\unskip}     \fi
\ifx \showISSN     \undefined \def \showISSN      #1{\unskip}     \fi
\ifx \showLCCN     \undefined \def \showLCCN      #1{\unskip}     \fi
\ifx \shownote     \undefined \def \shownote      #1{#1}          \fi
\ifx \showarticletitle \undefined \def \showarticletitle #1{#1}   \fi
\ifx \showURL      \undefined \def \showURL       {\relax}        \fi
% invisible to TeX
\providecommand\bibfield[2]{#2}
\providecommand\bibinfo[2]{#2}
\providecommand\natexlab[1]{#1}
\providecommand\showeprint[2][]{arXiv:#2}

\bibitem[\protect\citeauthoryear{Awiszus, Schubert, and Rosenhahn}{Awiszus
  et~al\mbox{.}}{2020}]%
        {awiszus2020toad}
\bibfield{author}{\bibinfo{person}{Maren Awiszus}, \bibinfo{person}{Frederik
  Schubert}, {and} \bibinfo{person}{Bodo Rosenhahn}.}
  \bibinfo{year}{2020}\natexlab{}.
\newblock \showarticletitle{TOAD-GAN: coherent style level generation from a
  single example}. In \bibinfo{booktitle}{\emph{Proceedings of the AAAI
  Conference on Artificial Intelligence and Interactive Digital
  Entertainment}}, Vol.~\bibinfo{volume}{16}. \bibinfo{pages}{10--16}.
\newblock


\bibitem[\protect\citeauthoryear{Capps and Schrum}{Capps and Schrum}{2021}]%
        {capps2021using}
\bibfield{author}{\bibinfo{person}{Benjamin Capps} {and} \bibinfo{person}{Jacob
  Schrum}.} \bibinfo{year}{2021}\natexlab{}.
\newblock \showarticletitle{Using Multiple Generative Adversarial Networks to
  Build Better-Connected Levels for Mega Man}.
\newblock \bibinfo{journal}{\emph{arXiv preprint arXiv:2102.00337}}
  (\bibinfo{year}{2021}).
\newblock


\bibitem[\protect\citeauthoryear{Compton and Mateas}{Compton and
  Mateas}{2006}]%
        {10.5555/3023108.3023128}
\bibfield{author}{\bibinfo{person}{Kate Compton} {and} \bibinfo{person}{Michael
  Mateas}.} \bibinfo{year}{2006}\natexlab{}.
\newblock \showarticletitle{Procedural Level Design for Platform Games}. In
  \bibinfo{booktitle}{\emph{Proceedings of the Second AAAI Conference on
  Artificial Intelligence and Interactive Digital Entertainment}} (Marina del
  Rey, California) \emph{(\bibinfo{series}{AIIDE'06})}.
  \bibinfo{publisher}{AAAI Press}, \bibinfo{pages}{109–111}.
\newblock


\bibitem[\protect\citeauthoryear{Dahlskog, Togelius, and Nelson}{Dahlskog
  et~al\mbox{.}}{2014}]%
        {dahlskog2014linear}
\bibfield{author}{\bibinfo{person}{Steve Dahlskog}, \bibinfo{person}{Julian
  Togelius}, {and} \bibinfo{person}{Mark~J Nelson}.}
  \bibinfo{year}{2014}\natexlab{}.
\newblock \showarticletitle{Linear levels through n-grams}. In
  \bibinfo{booktitle}{\emph{Proceedings of the 18th International Academic
  MindTrek Conference: Media Business, Management, Content \& Services}}.
  \bibinfo{pages}{200--206}.
\newblock


\bibitem[\protect\citeauthoryear{Guzdial, Reno, Chen, Smith, and Riedl}{Guzdial
  et~al\mbox{.}}{2018}]%
        {guzdial2018explainable}
\bibfield{author}{\bibinfo{person}{Matthew Guzdial}, \bibinfo{person}{Joshua
  Reno}, \bibinfo{person}{Jonathan Chen}, \bibinfo{person}{Gillian Smith},
  {and} \bibinfo{person}{Mark Riedl}.} \bibinfo{year}{2018}\natexlab{}.
\newblock \showarticletitle{Explainable PCGML via game design patterns}.
\newblock \bibinfo{journal}{\emph{arXiv preprint arXiv:1809.09419}}
  (\bibinfo{year}{2018}).
\newblock


\bibitem[\protect\citeauthoryear{Guzdial and Riedl}{Guzdial and Riedl}{2016}]%
        {guzdial2016toward}
\bibfield{author}{\bibinfo{person}{Matthew Guzdial} {and} \bibinfo{person}{Mark
  Riedl}.} \bibinfo{year}{2016}\natexlab{}.
\newblock \showarticletitle{Toward game level generation from gameplay videos}.
\newblock \bibinfo{journal}{\emph{arXiv preprint arXiv:1602.07721}}
  (\bibinfo{year}{2016}).
\newblock


\bibitem[\protect\citeauthoryear{Guzdial and Riedl}{Guzdial and Riedl}{2018}]%
        {guzdial2018automated}
\bibfield{author}{\bibinfo{person}{Matthew Guzdial} {and} \bibinfo{person}{Mark
  Riedl}.} \bibinfo{year}{2018}\natexlab{}.
\newblock \showarticletitle{Automated game design via conceptual expansion}. In
  \bibinfo{booktitle}{\emph{Proceedings of the AAAI Conference on Artificial
  Intelligence and Interactive Digital Entertainment}},
  Vol.~\bibinfo{volume}{14}.
\newblock


\bibitem[\protect\citeauthoryear{Jain, Isaksen, Holmg{\aa}rd, and
  Togelius}{Jain et~al\mbox{.}}{2016}]%
        {jain2016autoencoders}
\bibfield{author}{\bibinfo{person}{Rishabh Jain}, \bibinfo{person}{Aaron
  Isaksen}, \bibinfo{person}{Christoffer Holmg{\aa}rd}, {and}
  \bibinfo{person}{Julian Togelius}.} \bibinfo{year}{2016}\natexlab{}.
\newblock \showarticletitle{Autoencoders for level generation, repair, and
  recognition}. In \bibinfo{booktitle}{\emph{Proceedings of the ICCC Workshop
  on Computational Creativity and Games}}. \bibinfo{pages}{9}.
\newblock


\bibitem[\protect\citeauthoryear{Kalata}{Kalata}{2017}]%
        {kalata_2017}
\bibfield{author}{\bibinfo{person}{Kurt Kalata}.}
  \bibinfo{year}{2017}\natexlab{}.
\newblock \bibinfo{title}{Mega Man (Series Introduction)}.
\newblock
\newblock
\urldef\tempurl%
\url{http://www.hardcoregaming101.net/mega-man-series-introduction/}
\showURL{%
\tempurl}


\bibitem[\protect\citeauthoryear{Khalifa, Bontrager, Earle, and
  Togelius}{Khalifa et~al\mbox{.}}{2020}]%
        {khalifa2020pcgrl}
\bibfield{author}{\bibinfo{person}{Ahmed Khalifa}, \bibinfo{person}{Philip
  Bontrager}, \bibinfo{person}{Sam Earle}, {and} \bibinfo{person}{Julian
  Togelius}.} \bibinfo{year}{2020}\natexlab{}.
\newblock \showarticletitle{Pcgrl: Procedural content generation via
  reinforcement learning}. In \bibinfo{booktitle}{\emph{Proceedings of the AAAI
  Conference on Artificial Intelligence and Interactive Digital
  Entertainment}}, Vol.~\bibinfo{volume}{16}. \bibinfo{pages}{95--101}.
\newblock


\bibitem[\protect\citeauthoryear{Markov}{Markov}{1971}]%
        {markov1971extension}
\bibfield{author}{\bibinfo{person}{Andrey~Andreyevich Markov}.}
  \bibinfo{year}{1971}\natexlab{}.
\newblock \showarticletitle{Extension of the limit theorems of probability
  theory to a sum of variables connected in a chain}.
\newblock \bibinfo{journal}{\emph{Dynamic probabilistic systems}}
  \bibinfo{volume}{1} (\bibinfo{year}{1971}), \bibinfo{pages}{552--577}.
\newblock


\bibitem[\protect\citeauthoryear{Osborn, Summerville, and Mateas}{Osborn
  et~al\mbox{.}}{2017}]%
        {osborn2017automatic}
\bibfield{author}{\bibinfo{person}{Joseph Osborn}, \bibinfo{person}{Adam
  Summerville}, {and} \bibinfo{person}{Michael Mateas}.}
  \bibinfo{year}{2017}\natexlab{}.
\newblock \showarticletitle{Automatic mapping of nes games with mappy}. In
  \bibinfo{booktitle}{\emph{Proceedings of the 12th International Conference on
  the Foundations of Digital Games}}. \bibinfo{pages}{1--9}.
\newblock


\bibitem[\protect\citeauthoryear{Polikar}{Polikar}{2012}]%
        {polikar2012ensemble}
\bibfield{author}{\bibinfo{person}{Robi Polikar}.}
  \bibinfo{year}{2012}\natexlab{}.
\newblock \showarticletitle{Ensemble learning}.
\newblock In \bibinfo{booktitle}{\emph{Ensemble machine learning}}.
  \bibinfo{publisher}{Springer}, \bibinfo{pages}{1--34}.
\newblock


\bibitem[\protect\citeauthoryear{Sarkar and Cooper}{Sarkar and Cooper}{2020}]%
        {sarkar2020sequential}
\bibfield{author}{\bibinfo{person}{Anurag Sarkar} {and} \bibinfo{person}{Seth
  Cooper}.} \bibinfo{year}{2020}\natexlab{}.
\newblock \showarticletitle{Sequential segment-based level generation and
  blending using variational autoencoders}. In
  \bibinfo{booktitle}{\emph{International Conference on the Foundations of
  Digital Games}}. \bibinfo{pages}{1--9}.
\newblock


\bibitem[\protect\citeauthoryear{Sarkar and Cooper}{Sarkar and Cooper}{2021}]%
        {sarkar2021generating}
\bibfield{author}{\bibinfo{person}{Anurag Sarkar} {and} \bibinfo{person}{Seth
  Cooper}.} \bibinfo{year}{2021}\natexlab{}.
\newblock \showarticletitle{Generating and Blending Game Levels via
  Quality-Diversity in the Latent Space of a Variational Autoencoder}.
\newblock \bibinfo{journal}{\emph{arXiv preprint arXiv:2102.12463}}
  (\bibinfo{year}{2021}).
\newblock


\bibitem[\protect\citeauthoryear{Sarkar, Summerville, Snodgrass, Bentley, and
  Osborn}{Sarkar et~al\mbox{.}}{2020a}]%
        {sarkar2020exploring}
\bibfield{author}{\bibinfo{person}{Anurag Sarkar}, \bibinfo{person}{Adam
  Summerville}, \bibinfo{person}{Sam Snodgrass}, \bibinfo{person}{Gerard
  Bentley}, {and} \bibinfo{person}{Joseph Osborn}.}
  \bibinfo{year}{2020}\natexlab{a}.
\newblock \showarticletitle{Exploring level blending across platformers via
  paths and affordances}. In \bibinfo{booktitle}{\emph{Proceedings of the AAAI
  Conference on Artificial Intelligence and Interactive Digital
  Entertainment}}, Vol.~\bibinfo{volume}{16}. \bibinfo{pages}{280--286}.
\newblock


\bibitem[\protect\citeauthoryear{Sarkar, Yang, and Cooper}{Sarkar
  et~al\mbox{.}}{2020b}]%
        {sarkar2020conditional}
\bibfield{author}{\bibinfo{person}{Anurag Sarkar}, \bibinfo{person}{Zhihan
  Yang}, {and} \bibinfo{person}{Seth Cooper}.}
  \bibinfo{year}{2020}\natexlab{b}.
\newblock \showarticletitle{Conditional Level Generation and Game Blending}.
\newblock \bibinfo{journal}{\emph{arXiv preprint arXiv:2010.07735}}
  (\bibinfo{year}{2020}).
\newblock


\bibitem[\protect\citeauthoryear{Snodgrass and Ontan{\'o}n}{Snodgrass and
  Ontan{\'o}n}{2013}]%
        {snodgrass2013generating}
\bibfield{author}{\bibinfo{person}{Sam Snodgrass} {and}
  \bibinfo{person}{Santiago Ontan{\'o}n}.} \bibinfo{year}{2013}\natexlab{}.
\newblock \showarticletitle{Generating maps using markov chains}. In
  \bibinfo{booktitle}{\emph{Proceedings of the AAAI Conference on Artificial
  Intelligence and Interactive Digital Entertainment}},
  Vol.~\bibinfo{volume}{9}.
\newblock


\bibitem[\protect\citeauthoryear{Snodgrass and Onta{\~n}{\'o}n}{Snodgrass and
  Onta{\~n}{\'o}n}{2014a}]%
        {snodgrass2014experiments}
\bibfield{author}{\bibinfo{person}{Sam Snodgrass} {and}
  \bibinfo{person}{Santiago Onta{\~n}{\'o}n}.}
  \bibinfo{year}{2014}\natexlab{a}.
\newblock \showarticletitle{Experiments in map generation using Markov
  chains.}. In \bibinfo{booktitle}{\emph{Proceedings of the 9th International
  Conference on the Foundations of Digital Games}}.
\newblock


\bibitem[\protect\citeauthoryear{Snodgrass and Onta{\~n}{\'o}n}{Snodgrass and
  Onta{\~n}{\'o}n}{2014b}]%
        {snodgrass2014hierarchical}
\bibfield{author}{\bibinfo{person}{Sam Snodgrass} {and}
  \bibinfo{person}{Santiago Onta{\~n}{\'o}n}.}
  \bibinfo{year}{2014}\natexlab{b}.
\newblock \showarticletitle{A hierarchical approach to generating maps using
  markov chains}. In \bibinfo{booktitle}{\emph{Proceedings of the AAAI
  Conference on Artificial Intelligence and Interactive Digital
  Entertainment}}, Vol.~\bibinfo{volume}{10}.
\newblock


\bibitem[\protect\citeauthoryear{Snodgrass and Ontanon}{Snodgrass and
  Ontanon}{2015}]%
        {snodgrass2015hierarchical}
\bibfield{author}{\bibinfo{person}{Sam Snodgrass} {and}
  \bibinfo{person}{Santiago Ontanon}.} \bibinfo{year}{2015}\natexlab{}.
\newblock \showarticletitle{A hierarchical mdmc approach to 2d video game map
  generation}. In \bibinfo{booktitle}{\emph{Proceedings of the AAAI Conference
  on Artificial Intelligence and Interactive Digital Entertainment}},
  Vol.~\bibinfo{volume}{11}.
\newblock


\bibitem[\protect\citeauthoryear{Snodgrass and Onta{\~n}{\'o}n}{Snodgrass and
  Onta{\~n}{\'o}n}{2016}]%
        {snodgrass2016controllable}
\bibfield{author}{\bibinfo{person}{Sam Snodgrass} {and}
  \bibinfo{person}{Santiago Onta{\~n}{\'o}n}.} \bibinfo{year}{2016}\natexlab{}.
\newblock \showarticletitle{Controllable Procedural Content Generation via
  Constrained Multi-Dimensional Markov Chain Sampling.}. In
  \bibinfo{booktitle}{\emph{IJCAI}}. \bibinfo{pages}{780--786}.
\newblock


\bibitem[\protect\citeauthoryear{Snodgrass and Onta{\~n}{\'o}n}{Snodgrass and
  Onta{\~n}{\'o}n}{2017}]%
        {snodgrass2017Learning}
\bibfield{author}{\bibinfo{person}{Sam Snodgrass} {and}
  \bibinfo{person}{Santiago Onta{\~n}{\'o}n}.} \bibinfo{year}{2017}\natexlab{}.
\newblock \showarticletitle{Learning to Generate Video Game Maps Using Markov
  Models}.
\newblock \bibinfo{journal}{\emph{IEEE Transactions on Computational
  Intelligence and AI in Games}} \bibinfo{volume}{9}, \bibinfo{number}{4}
  (\bibinfo{year}{2017}), \bibinfo{pages}{410--422}.
\newblock
\urldef\tempurl%
\url{https://doi.org/10.1109/TCIAIG.2016.2623560}
\showDOI{\tempurl}


\bibitem[\protect\citeauthoryear{Snodgrass and Ontan{\'o}n}{Snodgrass and
  Ontan{\'o}n}{2017}]%
        {snodgrass2017procedural}
\bibfield{author}{\bibinfo{person}{Sam Snodgrass} {and}
  \bibinfo{person}{Santiago Ontan{\'o}n}.} \bibinfo{year}{2017}\natexlab{}.
\newblock \showarticletitle{Procedural level generation using multi-layer level
  representations with mdmcs}. In \bibinfo{booktitle}{\emph{2017 IEEE
  conference on computational intelligence and games (CIG)}}. IEEE,
  \bibinfo{pages}{280--287}.
\newblock


\bibitem[\protect\citeauthoryear{Summerville}{Summerville}{2018}]%
        {summerville2018expanding}
\bibfield{author}{\bibinfo{person}{Adam Summerville}.}
  \bibinfo{year}{2018}\natexlab{}.
\newblock \showarticletitle{Expanding expressive range: Evaluation
  methodologies for procedural content generation}. In
  \bibinfo{booktitle}{\emph{Proceedings of the AAAI Conference on Artificial
  Intelligence and Interactive Digital Entertainment}},
  Vol.~\bibinfo{volume}{14}.
\newblock


\bibitem[\protect\citeauthoryear{Summerville, Behrooz, Mateas, and
  Jhala}{Summerville et~al\mbox{.}}{2015a}]%
        {Summerville2015TheLO}
\bibfield{author}{\bibinfo{person}{A. Summerville}, \bibinfo{person}{M.
  Behrooz}, \bibinfo{person}{M. Mateas}, {and} \bibinfo{person}{A. Jhala}.}
  \bibinfo{year}{2015}\natexlab{a}.
\newblock \showarticletitle{The Learning of Zelda: Data-Driven Learning of
  Level Topology}.
\newblock


\bibitem[\protect\citeauthoryear{Summerville and Mateas}{Summerville and
  Mateas}{2016}]%
        {summerville2016super}
\bibfield{author}{\bibinfo{person}{Adam Summerville} {and}
  \bibinfo{person}{Michael Mateas}.} \bibinfo{year}{2016}\natexlab{}.
\newblock \showarticletitle{Super mario as a string: Platformer level
  generation via lstms}.
\newblock \bibinfo{journal}{\emph{arXiv preprint arXiv:1603.00930}}
  (\bibinfo{year}{2016}).
\newblock


\bibitem[\protect\citeauthoryear{Summerville, Philip, and Mateas}{Summerville
  et~al\mbox{.}}{2015b}]%
        {Summerville2015MCMCTSP4}
\bibfield{author}{\bibinfo{person}{A. Summerville}, \bibinfo{person}{Shweta
  Philip}, {and} \bibinfo{person}{M. Mateas}.}
  \bibinfo{year}{2015}\natexlab{b}.
\newblock \showarticletitle{MCMCTS PCG 4 SMB : Monte Carlo Tree Search to Guide
  Platformer Level Generation}.
\newblock


\bibitem[\protect\citeauthoryear{Summerville, Snodgrass, Guzdial, Holmg{\aa}rd,
  Hoover, Isaksen, Nealen, and Togelius}{Summerville et~al\mbox{.}}{2018}]%
        {summerville2018procedural}
\bibfield{author}{\bibinfo{person}{Adam Summerville}, \bibinfo{person}{Sam
  Snodgrass}, \bibinfo{person}{Matthew Guzdial}, \bibinfo{person}{Christoffer
  Holmg{\aa}rd}, \bibinfo{person}{Amy~K Hoover}, \bibinfo{person}{Aaron
  Isaksen}, \bibinfo{person}{Andy Nealen}, {and} \bibinfo{person}{Julian
  Togelius}.} \bibinfo{year}{2018}\natexlab{}.
\newblock \showarticletitle{Procedural content generation via machine learning
  (PCGML)}.
\newblock \bibinfo{journal}{\emph{IEEE Transactions on Games}}
  \bibinfo{volume}{10}, \bibinfo{number}{3} (\bibinfo{year}{2018}),
  \bibinfo{pages}{257--270}.
\newblock


\bibitem[\protect\citeauthoryear{Summerville, Snodgrass, Mateas, and {n}'{o}n
  Villar}{Summerville et~al\mbox{.}}{2016}]%
        {summerville2016vglc}
\bibfield{author}{\bibinfo{person}{Adam~James Summerville},
  \bibinfo{person}{Sam Snodgrass}, \bibinfo{person}{Michael Mateas}, {and}
  \bibinfo{person}{Santiago~Onta {n}'{o}n Villar}.}
  \bibinfo{year}{2016}\natexlab{}.
\newblock \showarticletitle{The VGLC: The Video Game Level Corpus}.
\newblock \bibinfo{journal}{\emph{Proceedings of the 7th Workshop on Procedural
  Content Generation}} (\bibinfo{year}{2016}).
\newblock


\bibitem[\protect\citeauthoryear{Torrado, Khalifa, Green, Justesen, Risi, and
  Togelius}{Torrado et~al\mbox{.}}{2020}]%
        {torrado2020bootstrapping}
\bibfield{author}{\bibinfo{person}{Ruben~Rodriguez Torrado},
  \bibinfo{person}{Ahmed Khalifa}, \bibinfo{person}{Michael~Cerny Green},
  \bibinfo{person}{Niels Justesen}, \bibinfo{person}{Sebastian Risi}, {and}
  \bibinfo{person}{Julian Togelius}.} \bibinfo{year}{2020}\natexlab{}.
\newblock \showarticletitle{Bootstrapping conditional gans for video game level
  generation}. In \bibinfo{booktitle}{\emph{2020 IEEE Conference on Games
  (CoG)}}. IEEE, \bibinfo{pages}{41--48}.
\newblock


\bibitem[\protect\citeauthoryear{Volz, Schrum, Liu, Lucas, Smith, and
  Risi}{Volz et~al\mbox{.}}{2018}]%
        {volz2018evolving}
\bibfield{author}{\bibinfo{person}{Vanessa Volz}, \bibinfo{person}{Jacob
  Schrum}, \bibinfo{person}{Jialin Liu}, \bibinfo{person}{Simon~M Lucas},
  \bibinfo{person}{Adam Smith}, {and} \bibinfo{person}{Sebastian Risi}.}
  \bibinfo{year}{2018}\natexlab{}.
\newblock \showarticletitle{Evolving mario levels in the latent space of a deep
  convolutional generative adversarial network}. In
  \bibinfo{booktitle}{\emph{Proceedings of the Genetic and Evolutionary
  Computation Conference}}. \bibinfo{pages}{221--228}.
\newblock


\bibitem[\protect\citeauthoryear{Yang, Sarkar, and Cooper}{Yang
  et~al\mbox{.}}{2020}]%
        {yang2020game}
\bibfield{author}{\bibinfo{person}{Zhihan Yang}, \bibinfo{person}{Anurag
  Sarkar}, {and} \bibinfo{person}{Seth Cooper}.}
  \bibinfo{year}{2020}\natexlab{}.
\newblock \showarticletitle{Game level clustering and generation using Gaussian
  mixture VAEs}. In \bibinfo{booktitle}{\emph{Proceedings of the AAAI
  Conference on Artificial Intelligence and Interactive Digital
  Entertainment}}, Vol.~\bibinfo{volume}{16}. \bibinfo{pages}{137--143}.
\newblock


\bibitem[\protect\citeauthoryear{{Zafar}, {Irfan}, and {Sabir}}{{Zafar}
  et~al\mbox{.}}{2019}]%
        {8974310}
\bibfield{author}{\bibinfo{person}{A. {Zafar}}, \bibinfo{person}{A. {Irfan}},
  {and} \bibinfo{person}{M.~Z. {Sabir}}.} \bibinfo{year}{2019}\natexlab{}.
\newblock \showarticletitle{Generating General Levels using Markov Chains}. In
  \bibinfo{booktitle}{\emph{2019 11th Computer Science and Electronic
  Engineering (CEEC)}}. \bibinfo{pages}{134--138}.
\newblock
\urldef\tempurl%
\url{https://doi.org/10.1109/CEEC47804.2019.8974310}
\showDOI{\tempurl}


\end{thebibliography}

\appendix

\end{document}